\documentclass[letterpaper, 10 pt, conference]{ieeeconf}  

\IEEEoverridecommandlockouts                              

\usepackage{cite}
\usepackage{amsmath,amssymb,amsfonts}
\usepackage{algorithmic}
\usepackage{graphicx}
\usepackage{textcomp}
\usepackage{xcolor}
\usepackage{dsfont}
\usepackage{xurl}
\usepackage{multirow}
\usepackage{svg}
\usepackage{tikz}

\usepackage{pifont}
\newcommand{\cmark}{\ding{51}}%
\newcommand{\xmark}{\ding{55}}%

\definecolor{ac_idle}{RGB}{31,119,180}
\definecolor{ac_approach}{RGB}{179, 198, 229}
\definecolor{ac_lift}{RGB}{245, 190, 130}
\definecolor{ac_hold}{RGB}{168, 221, 147}
\definecolor{ac_drink}{RGB}{213, 125, 190} 
\definecolor{ac_place}{RGB}{81, 158, 62} 
\definecolor{ac_retreat}{RGB}{239, 134, 54}
\definecolor{ac_pour}{RGB}{195, 60, 53}

\newcommand{\ColorBox}[1]{%
    \raisebox{0.5ex}{\colorbox{#1}{\hspace{0.05cm}\rule{0pt}{0.05cm}}}%
}

\title{\LARGE \bf
Real-Time Manipulation Action Recognition \\with a Factorized Graph Sequence Encoder\\
}

\author{Enes Erdogan$^{1}$, Eren Erdal Aksoy$^{2}$, and Sanem Sariel$^{1}$
\thanks{*This work has been submitted to the IEEE for possible publication. Copyright may be transferred without notice, after which this version may no longer be accessible}
\thanks{**This work was supported by the Scientific and Technological Research Council of Türkiye under Grant 119E-436.}
\thanks{$^{1}$Enes Erdogan and Sanem Sariel are with the Artificial Intelligence and Robotics Lab, Faculty of Computer
and Informatics Engineering, Istanbul Technical University, Istanbul, Turkey {\tt\small \{erdogane16, sariel\}@itu.edu.tr}}%
\thanks{$^{2}$Eren Erdal Aksoy is with the School of Information Technology, Halmstad University, Halmstad, Sweden
        {\tt\small eren.aksoy@hh.se}}%
}

\begin{document}

\maketitle
\thispagestyle{empty}
\pagestyle{empty}

\begin{abstract}

Recognition of human manipulation actions in real-time is essential for safe and effective human-robot interaction and collaboration. The challenge lies in developing a model that is both lightweight enough for real-time execution and capable of generalization. While some existing methods in the literature can run in real-time, they struggle with temporal scalability, i.e., they fail to adapt to long-duration manipulations effectively. To address this, leveraging the generalizable scene graph representations, we propose a new Factorized Graph Sequence Encoder network that not only runs in real-time but also scales effectively in the temporal dimension, thanks to its factorized encoder architecture. Additionally, we introduce  Hand Pooling operation, a simple pooling operation for more focused extraction of the graph-level embeddings. Our model outperforms the previous state-of-the-art real-time approach, achieving a 14.3\% and 5.6\% improvement in F1-macro score on the KIT Bimanual Action (Bimacs) Dataset and Collaborative Action (CoAx) Dataset, respectively. Moreover, we conduct an extensive ablation study to validate our network design choices. 
Finally, we compare our model with its architecturally similar RGB-based model on the Bimacs dataset and show the limitations of this model in contrast to ours on such an object-centric manipulation dataset.

\end{abstract}

\section{INTRODUCTION}

Manipulation action recognition is a crucial cognitive competence for robots, enabling them to autonomously detect and classify human activities from continuous sensory data, such as image streams. In contrast,  Real-Time Manipulation Recognition (RT-MR), a more advanced and demanding field, rather focuses on recognizing manipulation instances with minimal latency as soon as they emerge. The real-time aspect is of critical importance in scenarios where immediate responses are necessary, such as in \textit{Human-Robot Collaboration} (HRC), where humans manipulate objects to accomplish specific goals with robot assistance~\cite{Chandrasekaran15}.    

RT-MR models are required to be computationally efficient to ensure seamless operation with minimal latency. 
In the context of HRC, these recognition models are also expected to capture semantically rich abstract knowledge representation~\cite{dreher, sec, esec} to help robots gain more autonomy.    
However,  
high-dimensional raw RGB image streams not only lack the inherent semantic meaning of manipulations but also increase the workload for end-to-end training or inference.
Representing the scenes with graphs has emerged as a promising approach to address these challenges, offering an abstract yet comprehensive representation of the scene~\cite{dreher,sec,esec,gnet, 2ggcn, pgcn,h2o}.
These graphs are symbolic and capture the underlying semantic structure of the scene in a relatively lower dimensional space, thereby enabling real-time processing.

With this motivation, we, in this work, study the real-time recognition of human manipulation actions using symbolic scene graphs, where nodes represent objects and edges store semantic embeddings for spatial and/or temporal relations between objects, such as \textit{touching, being above, moving together,} and \textit{getting close}, among others. 
Most contemporary research in graph-based manipulation recognition models either omit real-time constraints~\cite{gnet, assign, 2ggcn, pgcn} or rely on a temporally concatenated graph representation that is infeasible to scale temporally~\cite{dreher, h2o}, limiting the model to recognize relatively long manipulation episodes. 
Therefore, to address these challenges, we introduce a new Factorized Graph Sequence Encoder network to recognize manipulation actions in real-time using the scene graph representation only. Inspired by the factorized encoder design in ViViT~\cite{vivit}, more specifically ViViT Model 2, our model separates spatiotemporal feature extraction into Graph Encoder and Sequence Encoder combined with a new Hand Pooling operation, improving parameter efficiency and scalability. 
Since the proposed model encodes each graph separately, it can dynamically scale along the temporal axis without any need to increase the depth of the graph neural network, in contrast to previous works \cite{dreher, h2o}. 
Our new parameter-free Hand Pooling operation selects node embeddings belonging to the hands, boosting the recognition performance.
Additionally, we employ a sliding window approach with the majority voting to improve recognition performance during the inference at a cost of short constant delay.

In contrast to real-time capable state-of-the-art manipulation models~\cite{h2o}, we achieve 14.3\% improvement in terms of F1 macro score on the KIT Bimanual Action (Bimacs) Dataset~\cite{dreher}. Furthermore, when allowing a slightly higher delay, our model achieves results comparable to offline models that process entire videos at once.  
Additional benchmarking on the Collaborative Action Dataset (CoAx)~\cite{coax} reveals that our proposed model outperforms the nearest competitor by 5.6\%, validating the generalizability of our approach.

The summary of our contributions is as follows:
\begin{itemize}
    \item 
    We introduce a new Factorized Graph Sequence Encoder combined with a new Hand Pooling operation that improves the F1 macro score by 14.3\%  and 5.6\% in comparison to the nearest competitor~\cite{h2o} on Bimacs~\cite{dreher} and CoAx~\cite{coax} datasets, respectively. 
    
    \item Addressing the limitations of previous approaches, we demonstrate that our network design supports temporal scalability, meaning that as the input sequence length increases, the model performs better.
\end{itemize}
 
We here note that in the context of action understanding, different terminologies exist in the literature. The term  \textit{``action"}  broadly encompasses any individual behavior, such as \textit{walking, jumping,} or \textit{pushing}. Since in this paper, our focus is specifically on human activities in HRC, involving \textit{hands interacting with objects} to accomplish a specific goal such as \textit{pushing a cup, cutting bread with a knife,} or \textit{stirring tea with a spoon}, we use a more precise term such as \textit{``manipulation"} or \textit{``manipulation action"}, which is also referred to as \textit{Human Object Interaction} in the literature. 
Therefore, we note that the literature on real-time action recognition~\cite{zhang2016real,Carlos24,Liu_2018} is not relevant to the field of RT-MR.

\section{RELATED WORKS}
Different from general-purpose RGB-based recognition models for common computer vision tasks~\cite{zhang2016real,Carlos24,Liu_2018,lstr, testra, oadtr}, our proposed work studies the recognition of human manipulation actions using scene graphs. Therefore, in the following sections, we review the graph-based manipulation recognition literature, dividing the subject into two subsections: offline and real-time models. 

\subsection{Offline Graph-Based Manipulation Recognition Models}
There exists a large corpus of work in graph-based scene representation for manipulation recognition~\cite{Sridhar08,Kjellstrom11,yang2013detection,Aksoy2010}. Most of these works, however, work offline in a batch mode.
For instance, Akyol et al.~\cite{gnet} propose a two-headed manipulation recognition and prediction network based on Variational Graph Autoencoders~\cite{kipf2016variational}, where reconstruction is not necessary. However, this work assumes that the key scene graphs are known prior. Also, the proposed model only accepts a single graph as an input, thus, the model lacks temporal understanding, making the model infeasible for real-time applications.

Morais et al.~\cite{assign} follow a different approach and model each entity in the scene with their state evolving throughout the video sparsely and asynchronously by interacting with each other. The state is a manipulation label for humans and an affordance label for objects. 
Node features are derived from low-level visual features extracted using a pre-trained Faster-RCNN~\cite{fasterRCNN} model and the messages between nodes are modeled as a type of attention, i.e., the cosine similarity between node features. 
This architecture is extended in~\cite{2ggcn} with a position-based object graph to improve its performance.

The work in~\cite{pgcn} proposes an encoder-decoder architecture for joint learning of both manipulation recognition and temporal segmentation tasks. 
Their contribution involves a novel attention-based graph convolution layer to encode scene graphs and a temporal pyramidal pooling module to decode these graph embeddings into framewise labels. Spatial position information is the only cue employed as node embedding to represent skeletons and objects in the scene. 
The edges are dynamically created between highly correlated nodes during manipulation, except for those between skeleton joints, which are defined naturally. 
Conventional 2D convolution operations are then applied to a generated $V\times T$ dimensional feature map, where $V$ is the number of objects and $T$ defines the length of the video. 
However, this design strictly assumes that the number of nodes throughout the video is constant, which is a highly restrictive assumption for complex manipulation sequences. Based on \cite{pgcn}, \cite{TFGCN} enhances temporal segmentation by introducing a Temporal Feature Fusion decoder while preserving feature space distances with Spectral Normalized Residual connections. However, the model in \cite{TFGCN} becomes 3.7 times larger than \cite{pgcn}, leading to higher computational complexity.

In contrast to these works, our model operates in real-time and does not rely on any prior knowledge about the number of graph nodes/edges, nor does it require low-level RGB features.

\subsection{Real-Time Graph-Based Manipulation Recognition Models}
\label{section:sota:online}
In the context of online manipulation recognition, Dreher et al. \cite{dreher} propose a model based on the graph encoder-decoder architecture~\cite{the_graphs}. They first extract graphs for each frame separately, using spatiotemporal relations introduced in~\cite{esec}. 
Next, in order to combine the sequential graphs, they introduce the temporal connections between the same nodes in consecutive graphs. 
However, considering that the graph neural networks are capable of propagating information to n-hop distant nodes where n denotes the number of layers, this design exhibits scalability limitations when the temporal length of the input increases. One might suggest that new layers could be added to compensate, but in return, over-smoothing~\cite{oversmoothing-analysis, oversmoothing-survey} might occur, which is a well-known phenomenon in deep graph networks, where no meaningful and distinguishable node embeddings are learned.

Another recent attempt~\cite{h2o} proposes a joint model for manipulation recognition and manipulation-conditioned motion forecasting, with a two-stage training. 
Initially, the manipulation action recognition module is trained, and subsequently, to predict the motion of the objects and hands, the model employs the predicted manipulation information in addition to the current graph sequence. In this graph representation, node embeddings consist of 3D object positions concatenated with one-hot encoded object categories. Furthermore, only edges between the hands and other objects are considered, where the edge feature is nothing but the distance between the hands and objects. As in the case of~\cite{dreher}, the consecutive graphs are linked with temporal edges. 
Consequently, the aforementioned criticisms regarding the limited temporal scalability of the model also apply to this study in~\cite{h2o}. 
Additionally, the discarding of edges between the objects may prevent the model from learning more complex manipulations.

The recent work in~\cite{mv_ohar_graph} employs skeleton data and applies the sliding window with a majority voting approach on top of the Spatial-Temporal Graph Convolutional Networks (ST-GCN)~\cite{stgcn}. The scalability issue is also valid for the ST-GCN model due to the temporal concatenation of sequential graphs. 

Our proposed model also differs from these real-time capable works due to our factorized encoder design, which enables temporal scaling of the network to enhance accuracy.

\section{METHOD}
In this section, we first explain the scene graph representation and then detail the proposed network architecture. Lastly, we discuss how the network is utilized in real-time using a sliding window approach coupled with majority voting.

\subsection{Graph Representation}
Given a human manipulation demonstration from a third-person view (see Fig.~\ref{fig:qualitative}), we follow the ontology of manipulation actions introduced in~\cite{TAMD13} and represent the scene with a graph, where nodes represent objects and edges denote spatiotemporal relations between the objects in the scene. 
As comprehensively defined in~\cite{esec}, there exist a total of 14 unique semantic relations, describing the static/spatial (such as above, below, inside, etc.)  and dynamic/temporal  (e.g., moving together, getting close, moving apart, etc.) aspects. These relations are derived from the 3D object bounding boxes in consecutive frames and form edge features to be stored in the scene graph.

Formally, streaming of a graph sequence can be defined as $S = \{G_0, G_1, ... G_t \}$, where $G_t$ represents the extracted scene graph at time step $t$. At any specific time $\tau$, the extracted graph is denoted as $G_\tau = (V_\tau, E_\tau)$, where $V_\tau$ is the set of nodes, expressed as $V_\tau = \{ v^{i}_\tau \}$. Each node $v^i_\tau$ is a one-hot-encoded object category.
Similarly, $E_\tau$ denotes the set of edges, given by $E_\tau = \{ e^{j}_\tau\}$, where each edge $e^{j}_\tau$ is represented as a 14-dimensional binary feature vector, i.e., $ e^{j}_\tau \in \{0, 1\}^{14}$.

\begin{figure}[!t]
\centerline{\includegraphics[width=0.5\textwidth]{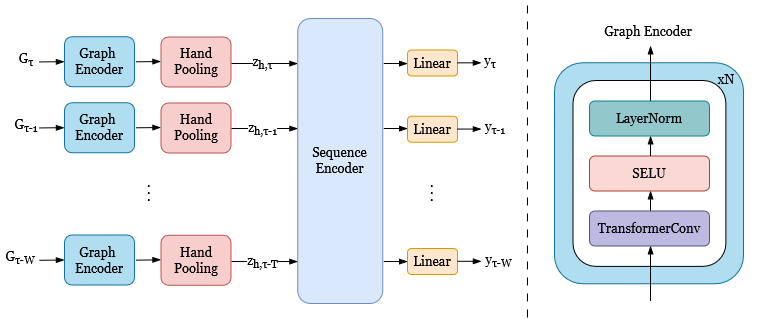}}
\caption{The proposed Factorized Graph Sequence Encoder (FGSE) network.}
\label{fig:model}
\end{figure}

\subsection{Factorized Graph Sequence Encoder (FGSE)  Network}

We propose a new Factorized Graph Sequence Encoder (FGSE) network to recognize manipulation actions in real-time from a stream of graph data. 
FGSE consists of two distinct encoder types: \textit{Graph Encoder} and \textit{Sequence Encoder}, combined with a new \textit{Hand Pooling} operation, as illustrated in Fig.~\ref{fig:model}.

The Graph Encoder (GE) utilizes a variant of a transformer-based graph convolutional layer called \textit{TransformerConv}~\cite{gtn}. It applies attention-based message-passing to refine the node embeddings, thus, each node aggregates information from the connected neighbor nodes by incorporating the edge features. 
This graph convolutional layer is further combined with the \textit{SELU} activation~\cite{selu} function and \textit{LayerNorm}~\cite{layernorm}, which is cascaded N times to finally construct the GE (see Fig.~\ref{fig:model}).

In manipulation action scenarios, hands are the \textit{main and only manipulators} interacting with the objects in the scene~\cite{TAMD13}. Therefore, it is reasonable to assume that hands accumulate more descriptive embeddings to infer types of performed manipulations. With this assumption, to obtain graph-level embeddings, we propose a simple and parameter-free operation, named Hand Pooling (HP), that selects node embeddings belonging to the hands in the initial graph. The combination of these two stages can be expressed as:

\begin{equation}
    HP(GE_\theta(G_\tau)) = \textbf{z}_{h,\tau}   ~~~, 
\end{equation}

where the Graph Encoder network, $GE$, is parametrized by $\theta$ and $\textbf{z}_{h,\tau}$ is the hand-corresponding ($h$) embedding vector pooled by $HP$ at time $\tau$ from the corresponding scene graph $G_\tau$.  

The Sequence Encoder (SE) is an encoder-only transformer~\cite{tr} that enables the model to learn temporal context by applying self-attention to hand embeddings (${\textbf{z}}_{h,\tau}$) in the input sequence. Finally, for each graph, a linear layer is applied to map those embeddings to manipulation labels. Stating these two layers combined formally:
   
\begin{equation}
    SE^L_\phi(\textbf{z}_{h,\tau-(W-1)}, \cdots, \textbf{z}_{h,\tau}) = (\hat{\textbf{y}}^{0}_{\tau-(W-1)}, \cdots, \hat{\textbf{y}}^{W-1}_\tau), 
\end{equation}

where Sequence Encoder network ($SE$) and linear layer ($L$), $SE^L$, is parametrized by $\phi$, and $W$ is the input window length of the model. The model prediction $\textbf{y}$ is the output vector of the \textit{Softmax} layer (which is omitted in the notation for the sake of simplicity), and its superscript denotes the relative position of the prediction within the given input. Note that, alternatively, the model could have predicted a single label for the whole input sequence by using the mean of the output embeddings or by employing, for instance, a class token. However, we observed that for the long sequences, this strategy significantly reduces the model's performance due to natural transitions among different types of manipulations throughout a long scenario. This is elaborated more in the discussion section.

The hallmark of our design is the separation of Graph and Sequence Encoders. This design allows the model to efficiently pass information among graphs even when the temporal length increases, regardless of the number of layers in the GE module. In addition, our new HP operation reduces the workload of SE by exclusively returning hand embeddings.

\subsection{Sliding Window with Majority Voting}

\begin{figure}[!t] 
\centerline{\includegraphics[width=0.35\textwidth]{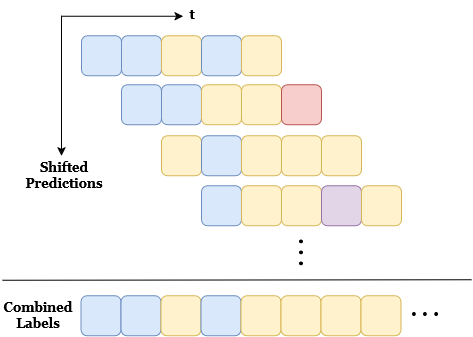}}
\caption{Majority voting to combine the shifted predictions. Starting from the top row, as the window slides along the temporal axis (x-axis), a new set of predictions is generated. 
The window size (W) is 5 in here. Each colored box represents a different predicted manipulation label.}
\label{fig:mv}
\end{figure}

The FGSE network returns a manipulation label for each corresponding input graph, as depicted in Fig.~\ref{fig:model}. During the inference process, we utilize a sliding window approach, which generates $W$ labels for a given graph, $G_\tau$. To combine these predictions, the majority voting algorithm is leveraged as illustrated in Fig.~\ref{fig:mv}. More formally, let $\hat{\textbf{y}}^w_\tau$ be the prediction vector, i.e., the output of the Softmax activation for $G_\tau$ as being the $w^{th}$ element in the sliding window. Thus, majority voting combines all predictions into a final one as:

\begin{equation}
    \tilde{\textbf{y}}_\tau=\arg\max_c \sum_{w=0}^{W-1} \mathds{1}_{\left(\arg\max(\hat{\textbf{y}}^w_\tau)=c\right)} ~~~,
\end{equation}

where $\tilde{\textbf{y}}_\tau$ is the combined labels at time $\tau$, and $\mathds{1}$ represents the indicator function.

As can be noticed, applying majority voting to a sliding window with a length of $W$ introduces a delay of $W/\text{FPS}$ seconds for the model output. 
Consequently, while a larger window enables the model to capture a richer local context, it comes with a cost of delay proportional to $W$. 

\section{EXPERIMENTS}

In this section, we present the experiment details, covering the network training process, the datasets used, the evaluation method, and a comparison of the results with state-of-the-art models from the literature.

\subsection{Training Setup}

We optimize the proposed FGSE network by minimizing the cross-entropy loss, averaged over the windows. Notably, majority voting is not applied during training. We experiment with varying window lengths, denoted as W followed by the respective value (e.g., W30 means window length of 30). Additionally, we observed that consecutive graphs are quite similar to each other unless the action changes. Therefore, in certain experiments, we downsampled the input sequence by a factor of 3 to accelerate training and testing without compromising accuracy, referring to this as D3. Note that during the metric calculations, we upsampled them back into the original scale for fair comparison.



Through empirical evaluation, we set the number of layers in both the Graph Encoder and Sequence Encoder to 2, i.e., the parameter N in Fig.~\ref{fig:model} is set to 2. 
Following the work in~\cite{dreher}, we doubled the Bimacs dataset (which will be detailed in the next section) by mirroring the graphs as a way of data augmentation. Further network and training parameter details can be found in the shared source code\footnote{\url{https://github.com/eneserdo/FGSE}}.

\subsection{Datasets}
We benchmark the proposed model on two publicly available datasets described below.

\textbf{KIT Bimanual Action (Bimacs) Dataset~\cite{dreher}} consists of 6 subjects performing 9 unique manipulation tasks relevant to kitchen and workshop environments, each repeated 10 times. In total, it comprises 2 hours and 18 minutes of recorded activities in RGB-D, covering 14 atomic manipulation categories. The manipulation samples are fully labeled for both hands individually and involve interactions with a total of 12 distinct objects. Additionally, the dataset provides the 3D object bounding box annotations for each object in the scene.

Bimacs provides the extracted scene graphs with the spatiotemporal relations in~\cite{esec}. Thus, we directly feed these graphs to our FGSE model. Note that since manipulations in Bimacs are labeled for each hand separately, we employ two linear layers to predict each manipulation performed by the left and right hands individually. 

\textbf{Collaborative Action (CoAx) Dataset~\cite{coax}} involves 6 subjects executing 3 industrial assembly manipulation tasks, one of which involves interaction with a collaborative robot. Each manipulation is repeated 10 times. The dataset contains a total of 1 hour and 58 minutes of RGB-D video data. The dataset features 10 distinct manipulation actions with 16 objects, and the frames are labeled as action-object pairs. Additionally, the dataset provides 3D object bounding boxes. Unlike Bimacs~\cite{dreher}, spatiotemporal relation information is not provided along with 3D object bounding boxes in CoAx. Following the work in~\cite{esec}, we compute these missing relations from the bounding boxes before the training.

In CoAx, even though there are 10 actions and 16 objects, which results in 160 possible label combinations, only 23 pairs are actually present. To simplify the model, we identify these existing pairs and merge them into a single label.

In both datasets, there might be noisy object detections and, consequently, incorrect relations between those objects. To mitigate this issue, we set an empirical threshold to filter out such relations. Specifically, if any two objects are too far apart, we remove the edge between them.

\subsection{Evaluation}

Macro and micro-averaged F1 scores are measured to report the success of each trained model. Following the work in~\cite{dreher}, we apply the leave-one-subject-out cross-validation approach to generate six folds for both datasets. 

\subsection{Results}

In this section, we present the results for two variants of our model with window lengths of 30 and 75. Table~\ref{tab:comparison} compares the recognition performance of our model with other relevant models on the Bimacs~\cite{dreher} dataset. We separate the benchmarked models based on their real-time capabilities, such as online versus offline models.

Among the online models (e.g., \cite{h2o} and \cite{dreher} in Table~\ref{tab:comparison}), our model (FGSE-W75-D3) achieves a significant improvement on the previous state-of-the-art model~\cite{h2o}, surpassing it by 14.3\% and 14.7\% in terms of F1-macro and F1-micro scores, respectively.

\begin{figure*}[!t]
\centerline{\includegraphics[width=1\textwidth]{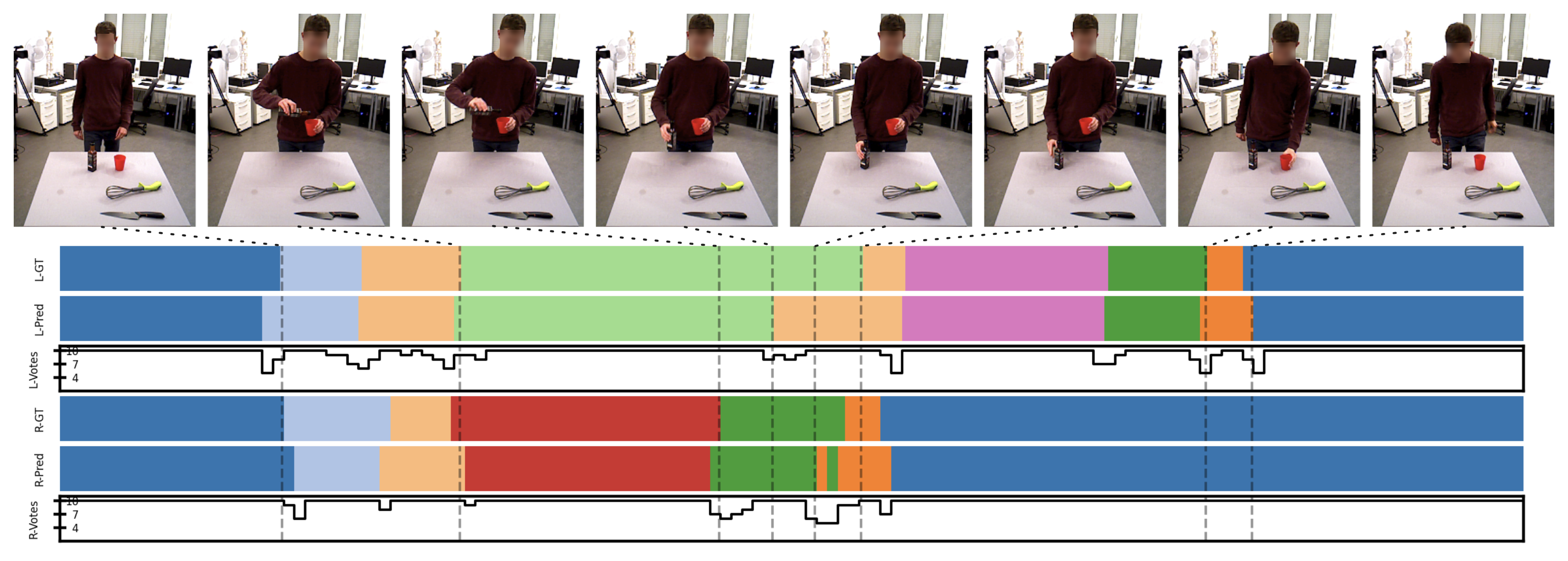}}
\caption{An example run from the test set of Bimacs \cite{dreher}, in which a person pours water from a bottle into a cup, and then drinks it. The top three rows show the ground-truth labels, predictions, and vote count in majority voting for the left hand, and the next three rows correspond to the right hand. Each color represents different actions: \ColorBox{ac_idle} idle, \ColorBox{ac_approach} approach, \ColorBox{ac_lift} lift, \ColorBox{ac_hold} hold, \ColorBox{ac_pour} pour, \ColorBox{ac_place} place, \ColorBox{ac_retreat} retreat, \ColorBox{ac_drink} drink.}
\label{fig:qualitative}
\end{figure*}

\begin{table}[!t] \centering
\caption{Manipulation Recognition Results on Bimacs~\cite{dreher}. }
\begin{tabular}{l|cc|cc} 
\hline
 Methods & Real-time & No visual & F1-  & F1- \\ 
 & capable & feature & macro &  micro\\ 
\hline
ASSIGN~\cite{assign, assign-supplementary} & \xmark & \xmark &  79.5 & 82.3 \\
PGCN~\cite{pgcn} & \xmark & \cmark & 81.5 & 86.9 \\
UQ-TFGCN~\cite{TFGCN} & \xmark & \cmark & 88.6 & 88.4 \\
\hline
Dreher et al.~\cite{dreher} & \cmark & \cmark &  63.0 & 64.0 \\
H2O+RGCN~\cite{h2o} & \cmark & \cmark & 66.0 & 68.0 \\
FGSE-W30-D3 (Ours) & \cmark & \cmark &  78.1 & 81.1 \\
FGSE-W75-D3 (Ours) & \cmark & \cmark &  80.3 & 82.7 \\
\hline
\end{tabular}
\label{tab:comparison}
\end{table}

Compared to the offline models (e.g., \cite{assign, assign-supplementary,pgcn} and \cite{TFGCN} in Table~\ref{tab:comparison}) that take the entire video at once, i.e., access the complete context and relations between the manipulations, our model (FGSE-W75-D3) achieves comparable results with \cite{assign, pgcn} in case of increasing the window length (W=75). We, however, note that the incorporation of visual features in \cite{assign} contradicts the original purpose of scene graphs. Scene graphs are designed to represent objects independently of their appearances or shapes, thereby making manipulation recognition more generalizable.

The offline model UQ-TFGCN~\cite{TFGCN} attains the highest performance among all models; however, it has the drawback of having the highest number of parameters (20.1M), which is 74 times more than our model. Similarly, PGCN~\cite{pgcn} has 21 times more parameters (5.4M) than our model.

Fig.~\ref{fig:qualitative} presents an illustrative sample from the Bimacs dataset and its qualitative analysis. In addition to the ground truth and the predicted labels for the sample video, we also included the vote counts in majority voting, which can be related to the confidence level of the model. In all predictions, our model demonstrates a high degree of confidence, except for instances involving transitions between distinct manipulations.

\begin{table}[!b]
\centering
\caption{Manipulation Recognition Results on CoAx~\cite{coax}. }
\begin{tabular}{l|cc}
\hline
Methods & F1-macro & F1-micro \\
\hline
Dreher et al.~\cite{dreher} & 60.0 & 70.0 \\
H2O+RGCN~\cite{h2o} & 87.0 & 90.0 \\
\hline
FGSE-W30 (Ours) & 90.7 & 92.8 \\
FGSE-W75-D3 (Ours) & \textbf{92.6} & \textbf{94.9} \\
\hline
\end{tabular}
\label{tab:coax}
\end{table}

Table~\ref{tab:coax} reports the obtained recognition results on the CoAx dataset~\cite{coax}. Our model yields a new state-of-the-art score, improving the nearest competitor~\cite{h2o} by 5.6\% in terms of F1-macro. Note that the results of Dreher et al.~\cite{dreher} on the CoAx dataset are taken from~\cite{h2o}.

A comparison of our model’s variants in both Table \ref{tab:comparison} and Table \ref{tab:coax}, FGSE-W75-D3 and FGSE-W30-D3, indicates that slightly relaxing the real-time constraints, i.e., increasing the window length, leads to improved performance by allowing the model to capture a larger local context. A further analysis on the impact of window length is given in the discussion section.

Regarding the runtime performance, with 269K parameters and 4.8 GFLOPS (on average), our proposed model FGSE achieves approximately 66 FPS on an Intel i9-12900K CPU with an NVIDIA GeForce RTX 3060 GPU, indicating that it is lightweight enough to run in real-time even on a low-end GPU card. 

To summarize, the results indicate that our model achieves a new state-of-the-art performance among real-time capable models. Moreover, it demonstrates promising performance even when compared to offline models, especially given its extremely parameter-efficient design relative to \cite{TFGCN} and \cite{pgcn}.

\section{DISCUSSION}

In this section, we pose various questions to better analyze the performance and characteristics of our model.

\textbf{How does performance change as the input window length increases?} 
We hypothesized that due to the factorized encoder design, our model would perform better at scaling in the temporal dimension compared to these models that concatenate the input graphs temporally~\cite{dreher, h2o}. 
As shown in Table~\ref{tab:winsize}, our experimental findings on Bimacs~\cite{dreher}  reveal that the performance of our model substantially improves when the window length is doubled from 10 to 20 graphs. 
After this particular point, although the performance continues to increase, the rate of improvement slows down, which can be interpreted as 20 graphs being sufficient to recognize most of the manipulations, and feeding in more graphs does not dramatically enhance recognition performance.

A similar improvement trend is also visible in the CoAx dataset~\cite{coax}. As indicated in the last row in Table~\ref{tab:winsize}, our model demonstrated an improvement of 9.1 points in terms of F1-macro when the number of graphs increased from 10 to 40. The results indicate that our network is better at scaling temporarily by design.

\begin{table}[!b] \centering
\caption{The F1-macro scores as window length increases on Bimacs~\cite{dreher}.}
\begin{tabular}{ll|cccc}
\hline
Dataset & Window length (W) & 10 & 20 & 30 & 40 \\
\hline
\multirow{ 3}{*}{Bimacs} & Dreher et al.~\cite{dreher} & 63.0 & 49.6 & 51.0 & N/A \\
& Dreher et al.~\cite{dreher} (scaled) & 63.0 & 51.2 & 42.9 & N/A \\
& FGSE (Ours) & 72.2 & 78.3 & 78.6 & 79.9 \\ 
\hline
CoAx & FGSE (Ours) & 83.1 & 87.9 & 90.7 & 92.2\\ 
\hline
\end{tabular}
\label{tab:winsize}
\end{table}

In this table, we also compare our model with a real-time capable model proposed by Dreher et al.~\cite{dreher} only, since the source code of H2O-RGCN~\cite{h2o} is yet to be publicly available. 
As mentioned in Section~\ref{section:sota:online}, the compared model in~\cite{dreher} constructs a single graph to represent a sequence of graphs by using temporal edges. This design becomes unscalable as the temporal length of the input grows, since graph neural networks can propagate information to nodes up to n hops away, where n is the number of layers. The first row in Table~\ref{tab:winsize} shows that the model's performance  in~\cite{dreher} worsened even though input data contains more information as the window length increases. While adding more layers could mitigate this issue, it may also lead to over-smoothing~\cite{oversmoothing-analysis, oversmoothing-survey}. To examine this, we doubled and tripled the number of processing steps in Dreher's model~\cite{dreher}, and as indicated in the second row, this approach also resulted in a similar performance failure. Note that the first three rows in Table~\ref{tab:winsize} only show results for the first fold due to high computational load in~\cite{dreher} during training.

\textbf{How does an RGB-only model perform?}
Considering the thrilling improvements in the RGB-based recognition models, a reasonable question might be how such a model would perform on the Bimacs dataset~\cite{dreher}. ViViT Model 2~\cite{vivit} was chosen for comparison due to architectural similarity, i.e., it has spatial and temporal encoders analogous to our Graph Encoder and Sequence Encoder. In this ViViT model, we used a pre-trained
spatial encoder and trained the temporal encoder from scratch. To make the comparison fair, we also employed a sliding window approach with majority voting during the test time. Due to high computational cost, in this experiment, we only performed tests with the first fold.

As shown in Table~\ref{tab:vivit}, the obtained F1 scores of the ViViT model are quite low compared to our proposed model. 
We believe that this underperformance is inherently related to the RGB-based approaches. As known, RGB-based models require a significant amount of training data to learn from high-dimensional raw image data. In our case, even though the network is partially pre-trained, the Bimacs dataset may not be sufficient for such a model, despite our effort to minimize the number of parameters in the temporal encoder part of the network. This poor generalization performance in small dataset settings makes the RGB-based models infeasible for HRC scenarios in which, for instance, a very specific cooking-related manipulation is supposed to be learned with very few data points to help the robot efficiently collaborate with the human user.

\begin{table}[!b]
\centering
\caption{Comparison with an RGB-only model (W30-D3) on Bimacs~\cite{dreher}.}
\begin{tabular}{l|cc}
\hline
Model & F1-macro & F1-micro \\
\hline
ViViT-Model 2~\cite{vivit} & 63.5 & 64.1  \\
FGSE (Ours) & \textbf{78.3} & \textbf{82.6} \\
\hline
\end{tabular}
\label{tab:vivit}
\end{table}

On the other hand, semantically rich symbolic scene graph-based methods are expected to be better at generalization from a few data points thanks to the very low dimensional representation space. 
In such a semantic representation, details irrelevant to the manipulation, such as varying light conditions and background clutter, are naturally disregarded, which might pose a significant challenge for RGB-based methods.
For instance, the same pouring manipulation executed with different objects (e.g., a cup versus a bottle) might be unrecognizable by the RGB-based model due to the shape and appearance changes of the objects in the scene. 
One might argue that the scene graphs also depend on an object recognition model, thus, it is nothing else than just shifting the burden of generalizability to the object detector. 
However, object detectors are particularly trained to identify objects with varying visual features, which makes them inherently more robust. 
Given that manipulation recognition is inherently object-centric, where the temporal relationships between objects and their environment matter more than instance-specific properties like object appearance or geometry, it is reasonable to break down the manipulation recognition task into object detection and graph-based recognition steps.

To conclude, our experimental findings in Table~\ref{tab:vivit} reveal that RGB-only models underperform on object-centric manipulation datasets with a limited number of samples, such as Bimacs. This observation underscores the limitations of such models in HRC scenarios.

\textbf{How Much Does Majority Voting Contribute?}
As an ablation study, we measure the impact of majority voting. As a first alternative, we take the average of final embeddings after the Sequence Encoder and use a single linear layer to predict a label that corresponds to the last graph in the window. More formally, the Sequence Encoder combined with linear layer predicts as: 

\begin{equation}
    SE^L_\phi(\textbf{z}_{h,\tau-(W-1)}, \cdots, \textbf{z}_{h,\tau}) = \hat{\textbf{y}}_\tau~~~~.   
\end{equation}



As a second alternative approach, we only use the label at the window's center, i.e., $\tilde{\textbf{y}}_\tau = \hat{\textbf{y}}^{W/2}_\tau$, without altering the loss function or applying the majority voting. The results in Table~\ref{tab:majvot} indicate that majority voting strongly improves the model's performance compared to these alternatives. 


\begin{table}[!t]
\centering
\caption{Impact of the sliding window with majority voting on Bimacs \cite{dreher} (D3).}
\begin{tabular}{l|cc}
\hline
Methods & F1-macro & F1-micro \\
\hline
Center of window & 76.9 & 79.2 \\
Single Pred. & 70.0 & 73.2 \\
\hline
Majority voting & \textbf{78.1} & \textbf{81.1} \\
\hline
\end{tabular}
\label{tab:majvot}
\end{table}

\textbf{Is Hand Pooling better than other pooling methods?}
To show the effectiveness of our new Hand Pooling operation, we compare it with widely used alternative methods in the literature. As shown in Table~\ref{tab:pooling}, Hand Pooling is considerably more effective than the naive global mean pooling method. It also outperforms the more sophisticated pooling operations such as Top-k pooling \cite{topkpooling} and SAGPool \cite{sagpooling}. 
Note that our Hand Pooling operation also has the advantage of having zero cost compared to other pooling methods.

\begin{table}[!b]
\centering
\caption{The comparison with alternative pooling methods (W30-D3).}
\begin{tabular}{l|cc}
\hline
Methods & F1-macro & F1-micro \\
\hline
Global mean pooling & 75.6 & 79.5 \\
Top-k pooling \cite{topkpooling} & 77.1 & 80.7 \\
SAGPool \cite{sagpooling} & 75.5 & 79.7 \\
\hline
Hand-Pooling (Ours) & \textbf{78.1} & \textbf{81.1} \\
\hline
\end{tabular}
\label{tab:pooling}
\end{table}

\textbf{What is the contribution of the Sequence Encoder?}
To quantify the contribution of the Sequence Encoder in our model, we compare it with classical recurrent architectures. 
The first row of Table~\ref{tab:seqenc} shows that removing the Sequence Encoder completely and relying merely on the Graph Encoder drops the performance significantly. 
This clearly reveals that the Sequence Encoder contributes to the performance by extracting the local context information.    
On the other hand, LSTM-based approaches could not reach the performance of the Encoder-only Transformer network, although bidirectional LSTM shows promising results.

\begin{table}[!t]
\centering
\caption{The comparison with alternatives of Sequence Encoder on Bimacs \cite{dreher} (W30-D3).}
\begin{tabular}{l|cc}
\hline
Seq. Enc. Variants & F1-macro & F1-micro \\
\hline
No Encoder & 61.2 & 69.1 \\
LSTM  & 69.4 & 74.1 \\
BiLSTM  & 77.1 & 80.7 \\
\hline
Encoder-only Transformer & \textbf{78.1} & \textbf{81.1} \\
\hline
\end{tabular}
\label{tab:seqenc}
\end{table}

\section{CONCLUSION}

In this work, we introduced a new Factorized Graph Sequence Encoder as a real-time manipulation action recognition model that effectively balances computational efficiency and temporal scalability. By leveraging scene graph representations and the factorized encoder architecture, our approach achieves state-of-the-art performance on the Bimacs and CoAx datasets, surpassing the previous best real-time model by 14.3\% and 5.6\%, respectively. An extensive ablation study validates our design choices and demonstrates the model’s temporal scalability. Additionally, to the best of our knowledge, we are the first to benchmark and demonstrate the limited capacity of an RGB-based model (ViViT Model 2) on object-centric Bimacs manipulation datasets.

As a future research direction, we plan to apply our Factorized Graph Sequence Encoder network to the skeleton-based human whole-body manipulation tasks. Furthermore, despite our model’s strong performance, we observe that extracting scene graphs from RGB-D data is quite noisy, primarily due to the unreliable performance of depth cameras and the limitations of the object detection models. To address these challenges, we plan to incorporate the confidence of the estimation into our model.

By addressing these challenges, we hope to further advance human manipulation recognition models, making them more adaptable to real-world scenarios and facilitating seamless human-robot collaboration.


\section*{ACKNOWLEDGMENT}
Computing resources used in this work were provided by the National Center for High Performance Computing of Turkey (UHeM) under grant number 4019762024. 

\bibliographystyle{IEEEtran}
\bibliography{references}

\begin{thebibliography}{10}
\providecommand{\url}[1]{#1}
\csname url@samestyle\endcsname
\providecommand{\newblock}{\relax}
\providecommand{\bibinfo}[2]{#2}
\providecommand{\BIBentrySTDinterwordspacing}{\spaceskip=0pt\relax}
\providecommand{\BIBentryALTinterwordstretchfactor}{4}
\providecommand{\BIBentryALTinterwordspacing}{\spaceskip=\fontdimen2\font plus
\BIBentryALTinterwordstretchfactor\fontdimen3\font minus \fontdimen4\font\relax}
\providecommand{\BIBforeignlanguage}[2]{{%
\expandafter\ifx\csname l@#1\endcsname\relax
\typeout{** WARNING: IEEEtran.bst: No hyphenation pattern has been}%
\typeout{** loaded for the language `#1'. Using the pattern for}%
\typeout{** the default language instead.}%
\else
\language=\csname l@#1\endcsname
\fi
#2}}
\providecommand{\BIBdecl}{\relax}
\BIBdecl

\bibitem{Chandrasekaran15}
B.~Chandrasekaran and J.~M. Conrad, ``Human-robot collaboration: A survey,'' in \emph{SoutheastCon}, 2015, pp. 1--8.

\bibitem{dreher}
C.~R.~G. Dreher, M.~Wächter, and T.~Asfour, ``Learning object-action relations from bimanual human demonstration using graph networks,'' \emph{IEEE Robotics and Automation Letters}, vol.~5, no.~1, pp. 187--194, 2020.

\bibitem{sec}
E.~E. Aksoy, A.~Orhan, and F.~W{\"o}rg{\"o}tter, ``Semantic decomposition and recognition of long and complex manipulation action sequences,'' \emph{International Journal of Computer Vision}, vol. 122, pp. 84--115, 2017.

\bibitem{esec}
\BIBentryALTinterwordspacing
F.~Ziaeetabar, T.~Kulvicius, M.~Tamosiunaite, and F.~Wörgötter, ``Recognition and prediction of manipulation actions using enriched semantic event chains,'' \emph{Robotics and Autonomous Systems}, vol. 110, pp. 173--188, 2018. [Online]. Available: \url{https://www.sciencedirect.com/science/article/pii/S0921889018303725}
\BIBentrySTDinterwordspacing

\bibitem{gnet}
G.~Akyol, S.~Sariel, and E.~E. Aksoy, ``A variational graph autoencoder for manipulation action recognition and prediction,'' in \emph{20th International Conference on Advanced Robotics (ICAR)}.\hskip 1em plus 0.5em minus 0.4em\relax IEEE, 2021, pp. 968--973.

\bibitem{2ggcn}
T.~Qiao, Q.~Men, F.~W.~B. Li, Y.~Kubotani, S.~Morishima, and H.~P.~H. Shum, ``Geometric features informed multi-person human-object interaction recognition in videos,'' in \emph{European Conference on Computer Vision (ECCV)}, 2022.

\bibitem{pgcn}
H.~Xing and D.~Burschka, ``Understanding spatio-temporal relations in human-object interaction using pyramid graph convolutional network,'' in \emph{IEEE/RSJ International Conference on Intelligent Robots and Systems (IROS)}, 2022, pp. 5195--5201.

\bibitem{h2o}
D.~Lagamtzis, F.~Schmidt, J.~Seyler, T.~Dang, and S.~Schober, ``Exploiting spatio-temporal human-object relations using graph neural networks for human action recognition and 3d motion forecasting,'' in \emph{IEEE/RSJ International Conference on Intelligent Robots and Systems (IROS)}, 2023, pp. 7832--7838.

\bibitem{assign}
R.~Morais, V.~Le, S.~Venkatesh, and T.~Tran, ``Learning asynchronous and sparse human-object interaction in videos,'' in \emph{Proceedings of the IEEE/CVF Conference on Computer Vision and Pattern Recognition (CVPR)}, June 2021, pp. 16\,041--16\,050.

\bibitem{vivit}
A.~Arnab, M.~Dehghani, G.~Heigold, C.~Sun, M.~Lu{\v{c}}i{\'c}, and C.~Schmid, ``Vivit: A video vision transformer,'' in \emph{International Conference on Computer Vision (ICCV)}, 2021.

\bibitem{coax}
D.~Lagamtzis, F.~Schmidt, J.~R. Seyler, and T.~Dang, ``Coax: Collaborative action dataset for human motion forecasting in an industrial workspace.'' in \emph{ICAART (3)}, 2022, pp. 98--105.

\bibitem{zhang2016real}
B.~Zhang, L.~Wang, Z.~Wang, Y.~Qiao, and H.~Wang, ``Real-time action recognition with enhanced motion vector cnns,'' in \emph{Proceedings of the IEEE conference on computer vision and pattern recognition}, 2016, pp. 2718--2726.

\bibitem{Carlos24}
A.~C. Cob-Parro, C.~Losada-Guti{\'e}rrez, M.~Marr{\'o}n-Romera, A.~Gardel-Vicente, and I.~Bravo-Mu{\~n}oz, ``A new framework for deep learning video based human action recognition on the edge,'' \emph{Expert Systems with Applications}, vol. 238, p. 122220, 2024.

\bibitem{Liu_2018}
\BIBentryALTinterwordspacing
K.~Liu, W.~Liu, C.~Gan, M.~Tan, and H.~Ma, ``T-c3d: Temporal convolutional 3d network for real-time action recognition,'' \emph{Proceedings of the AAAI Conference on Artificial Intelligence}, vol.~32, no.~1, 2018. [Online]. Available: \url{https://ojs.aaai.org/index.php/AAAI/article/view/12333}
\BIBentrySTDinterwordspacing

\bibitem{lstr}
M.~Xu, Y.~Xiong, H.~Chen, X.~Li, W.~Xia, Z.~Tu, and S.~Soatto, ``Long short-term transformer for online action detection,'' in \emph{Conference on Neural Information Processing Systems (NeurIPS)}, 2021.

\bibitem{testra}
Y.~Zhao and P.~Kr{\"a}henb{\"u}hl, ``Real-time online video detection with temporal smoothing transformers,'' in \emph{European Conference on Computer Vision (ECCV)}, 2022.

\bibitem{oadtr}
X.~Wang, S.~Zhang, Z.~Qing, Y.~Shao, Z.~Zuo, C.~Gao, and N.~Sang, ``Oadtr: Online action detection with transformers,'' in \emph{Proceedings of the IEEE/CVF International Conference on Computer Vision}, 2021, pp. 7565--7575.

\bibitem{Sridhar08}
M.~Sridhar, G.~A. Cohn, and D.~Hogg, ``Learning functional object-categories from a relational spatio-temporal representation,'' in \emph{Proc. 18th European Conference on Artificial Intelligence}, 2008, pp. 606--610.

\bibitem{Kjellstrom11}
H.~Kjellstr\"{o}m, J.~Romero, and D.~Kragi\'{c}, ``Visual object-action recognition: Inferring object affordances from human demonstration,'' \emph{Comput. Vis. Image Underst.}, vol. 115, no.~1, pp. 81--90, jan 2011.

\bibitem{yang2013detection}
Y.~Yang, C.~Ferm\"{u}ller, and Y.~Aloimonos, ``Detection of manipulation action consequences (mac),'' in \emph{Proceedings of the IEEE Conference on Computer Vision and Pattern Recognition}, 2013, pp. 2563--2570.

\bibitem{Aksoy2010}
E.~E. Aksoy, A.~Abramov, F.~W\"org\"otter, and B.~Dellen, ``Categorizing object-action relations from semantic scene graphs,'' in \emph{IEEE International Conference on Robotics and Automation (ICRA)}, may 2010, pp. 398--405.

\bibitem{kipf2016variational}
T.~N. Kipf and M.~Welling, ``Variational graph auto-encoders,'' \emph{arXiv preprint arXiv:1611.07308}, 2016.

\bibitem{fasterRCNN}
S.~Ren, K.~He, R.~Girshick, and J.~Sun, ``Faster r-cnn: Towards real-time object detection with region proposal networks,'' \emph{IEEE transactions on pattern analysis and machine intelligence}, vol.~39, no.~6, pp. 1137--1149, 2016.

\bibitem{TFGCN}
H.~Xing and D.~Burschka, ``Understanding human activity with uncertainty measure for novelty in graph convolutional networks,'' \emph{The International Journal of Robotics Research}, p. 02783649241287800, 2024.

\bibitem{the_graphs}
\BIBentryALTinterwordspacing
P.~Battaglia, J.~B.~C. Hamrick, V.~Bapst, A.~Sanchez, V.~Zambaldi, M.~Malinowski, A.~Tacchetti, D.~Raposo, A.~Santoro, R.~Faulkner, C.~Gulcehre, F.~Song, A.~Ballard, J.~Gilmer, G.~E. Dahl, A.~Vaswani, K.~Allen, C.~Nash, V.~J. Langston, C.~Dyer, N.~Heess, D.~Wierstra, P.~Kohli, M.~Botvinick, O.~Vinyals, Y.~Li, and R.~Pascanu, ``Relational inductive biases, deep learning, and graph networks,'' \emph{arXiv}, 2018. [Online]. Available: \url{https://arxiv.org/pdf/1806.01261.pdf}
\BIBentrySTDinterwordspacing

\bibitem{oversmoothing-analysis}
N.~Keriven, ``Not too little, not too much: a theoretical analysis of graph (over) smoothing,'' \emph{Advances in Neural Information Processing Systems}, vol.~35, pp. 2268--2281, 2022.

\bibitem{oversmoothing-survey}
T.~K. Rusch, M.~M. Bronstein, and S.~Mishra, ``A survey on oversmoothing in graph neural networks,'' \emph{arXiv preprint arXiv:2303.10993}, 2023.

\bibitem{mv_ohar_graph}
\BIBentryALTinterwordspacing
M.~Dallel, V.~Havard, Y.~Dupuis, and D.~Baudry, ``A sliding window based approach with majority voting for online human action recognition using spatial temporal graph convolutional neural networks,'' in \emph{Proceedings of the 2022 7th International Conference on Machine Learning Technologies}, ser. ICMLT '22.\hskip 1em plus 0.5em minus 0.4em\relax New York, NY, USA: Association for Computing Machinery, 2022, p. 155–163. [Online]. Available: \url{https://doi.org/10.1145/3529399.3529425}
\BIBentrySTDinterwordspacing

\bibitem{stgcn}
S.~Yan, Y.~Xiong, and D.~Lin, ``Spatial temporal graph convolutional networks for skeleton-based action recognition,'' in \emph{Proceedings of the Thirty-Second AAAI Conference on Artificial Intelligence and Thirtieth Innovative Applications of Artificial Intelligence Conference and Eighth AAAI Symposium on Educational Advances in Artificial Intelligence}, ser. AAAI'18/IAAI'18/EAAI'18.\hskip 1em plus 0.5em minus 0.4em\relax AAAI Press, 2018.

\bibitem{TAMD13}
F.~W\"org\"otter, E.~E. Aksoy, N.~Kr\"uger, J.~Piater, A.~Ude, and M.~Tamosiunaite, ``A simple ontology of manipulation actions based on hand-object relations,'' \emph{IEEE Transactions on Autonomous Mental Development}, vol.~5, no.~2, pp. 117--134, 2013.

\bibitem{gtn}
\BIBentryALTinterwordspacing
Y.~Shi, Z.~Huang, S.~Feng, H.~Zhong, W.~Wang, and Y.~Sun, ``Masked label prediction: Unified message passing model for semi-supervised classification,'' in \emph{Proceedings of the Thirtieth International Joint Conference on Artificial Intelligence, {IJCAI-21}}, Z.-H. Zhou, Ed.\hskip 1em plus 0.5em minus 0.4em\relax International Joint Conferences on Artificial Intelligence Organization, 8 2021, pp. 1548--1554, main Track. [Online]. Available: \url{https://doi.org/10.24963/ijcai.2021/214}
\BIBentrySTDinterwordspacing

\bibitem{selu}
G.~Klambauer, T.~Unterthiner, A.~Mayr, and S.~Hochreiter, ``Self-normalizing neural networks,'' \emph{Advances in neural information processing systems}, vol.~30, 2017.

\bibitem{layernorm}
J.~L. Ba, J.~R. Kiros, and G.~E. Hinton, ``Layer normalization,'' \emph{arXiv preprint arXiv:1607.06450}, 2016.

\bibitem{tr}
A.~Vaswani, ``Attention is all you need,'' \emph{Advances in Neural Information Processing Systems}, 2017.

\bibitem{assign-supplementary}
R.~Morais, V.~Le, S.~Venkatesh, and T.~Tran, ``Learning asynchronous and sparse human-object interaction in videos-supplementary material.''

\bibitem{topkpooling}
H.~Gao and S.~Ji, ``Graph u-nets,'' in \emph{international conference on machine learning}.\hskip 1em plus 0.5em minus 0.4em\relax PMLR, 2019, pp. 2083--2092.

\bibitem{sagpooling}
J.~Lee, I.~Lee, and J.~Kang, ``Self-attention graph pooling,'' in \emph{International conference on machine learning}.\hskip 1em plus 0.5em minus 0.4em\relax pmlr, 2019, pp. 3734--3743.

\end{thebibliography}

\end{document}